\documentclass[runningheads]{llncs}
\usepackage[T1]{fontenc}
\usepackage[table,dvipsnames]{xcolor}
\usepackage{graphicx,verbatim}
\usepackage{amsmath,amssymb,amsfonts}
\usepackage{colortbl}
\usepackage{hyperref}
\hypersetup{
    colorlinks=true,
    linkcolor=blue,
    citecolor=blue,
    urlcolor=blue
}
\definecolor{tmi_blue}{cmyk}{0.39,0.37,0.0,0.15}
\usepackage{graphicx}
\usepackage{booktabs}
\usepackage{adjustbox}
\usepackage{multirow}
\usepackage{array}
\begin{document}
\title{MedNNS: Supernet-based Medical Task-Adaptive Neural Network Search}

\author{Lotfi Abdelkrim Mecharbat\inst{1} \and
Ibrahim Almakky \inst{1}
\and
Martin Takac \inst{1}
\and
Mohammad Yaqub \inst{1}}
\authorrunning{F. Author et al.}
%
\institute{Mohammed Bin Zayed University of Artificial Intelligence, Abu Dhabi, UAE
\email{firstname.lastname@mbzuai.ac.ae}}

\maketitle              
\begin{abstract}
Deep learning (DL) has achieved remarkable progress in the field of medical imaging. However, adapting DL models to medical tasks remains a significant challenge, primarily due to two key factors: (1) architecture selection, as different tasks necessitate specialized model designs, and (2) weight initialization, which directly impacts the convergence speed and final performance of the models. Although transfer learning from ImageNet is a widely adopted strategy, its effectiveness is constrained by the substantial differences between natural and medical images. To address these challenges, we introduce Medical Neural Network Search (MedNNS), the first Neural Network Search framework for medical imaging applications. MedNNS jointly optimizes architecture selection and weight initialization by constructing a meta-space that encodes datasets and models based on how well they perform together. We build this space using a Supernetwork-based approach, expanding the model zoo size by \textbf{51x times} over previous state-of-the-art (SOTA) methods. Moreover, we introduce rank loss and Fréchet Inception Distance (FID) loss into the construction of the space to capture inter-model and inter-dataset relationships, thereby achieving more accurate alignment in the meta-space. Experimental results across multiple datasets demonstrate that MedNNS significantly outperforms both ImageNet pretrained DL models and SOTA Neural Architecture Search (NAS) methods, achieving an average accuracy improvement of \textbf{1.7\%} across datasets while converging substantially faster. The code and the processed meta-space is available at \url{https://github.com/BioMedIA-MBZUAI/MedNNS}

\end{abstract}
\section{Introduction}
\label{sec:introduction}

Deep learning (DL) is transforming medical imaging in many aspects, such as enabling tumor type identification \cite{kouli2022automated}, skin lesion analysis \cite{pattnayak2024automated}, and diabetic retinopathy diagnosis \cite{qureshi2019recent}. These advances boost diagnostic accuracy and clinical efficiency, ultimately striving to improve patient outcomes. However, adapting DL techniques to medical tasks presents two fundamental challenges. First, choosing the optimal architecture, where different tasks often require different model designs tailored to capture subtle, domain-specific features \cite{raghu2019transfusion,godasu2020transfer}. Second, weight initialization strategies, which influence convergence during training, ultimately affecting the overall performance of the model \cite{boulila2024effective,skorski2021revisiting}.

\begin{figure}[t]
    \centering
    \includegraphics[width=1.\linewidth]{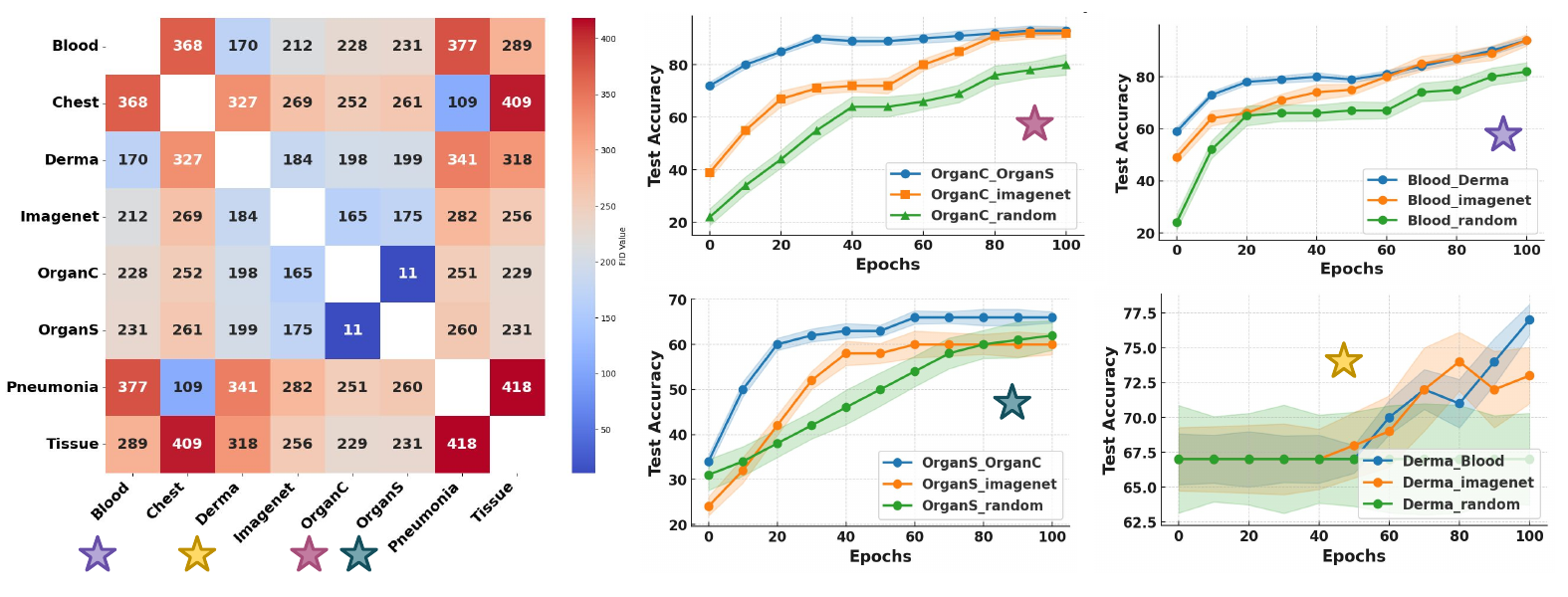}
    \caption{ (Left) Heatmap showing pairwise FID distance between MedMNIST datasets and ImageNet. (Right) Training curves for ResNet-18 using random, ImageNet, and nearest-FID dataset pertaining.}
    \label{fig:fid-finetunning}
\end{figure}

Transfer learning enhances convergence by providing a better initialization from models pretrained on large datasets like ImageNet \cite{deng2009imagenet}. However, medical images differ substantially from natural images in format, modality, and appearance \cite{prevedello2019challenges,raghu2019transfusion}, making feature reuse from ImageNet often ineffective \cite{kim2022transfer}. Investigating this, we compute the Fréchet Inception Distance (FID) \cite{yu2021frechet} between several MedMNIST \cite{yang2023medmnist} datasets and ImageNet \cite{deng2009imagenet} as shown in Fig. \ref{fig:fid-finetunning} and observe that every medical dataset has at least one other medical dataset that is closer in FID compared to ImageNet. Moreover, training ResNet-18 \cite{he2016deep} on four datasets with different initializations (Fig. \ref{fig:fid-finetunning}) demonstrates that pretraining on a more similar dataset accelerates convergence and improves the final performance, suggesting more effective feature reuse.

While selecting a more relevant dataset for pretraining improves feature transfer, the architecture itself plays an important role in determining which features are extracted and reused effectively. This is because the architecture’s inductive bias, which are the inherent assumptions it encodes about the data and task, dictates which features are prioritized during learning \cite{matsoukas2022makes}. This has led to growing interest in Neural Architecture Search (NAS), an automated approach that explores a space of architectures to find configurations that best align with task and data-specific characteristics. For a given dataset \( D \), NAS aims to discover an optimal architecture \( \alpha^* \) that minimizes the validation loss \( \mathcal{L}^{(D)}_{val} \) when evaluated on the model \( f_\alpha \) with its optimal weights \( w^*_\alpha \). Formally:

\begin{equation}
\alpha^* = \arg\min_{\alpha \in \mathcal{A}} \mathcal{L}^{(D)}_{val}\left(f_\alpha\left(w^*_\alpha\right)\right), \quad
where : \quad
w^*_\alpha = \arg\min_w \mathcal{L}^{(D)}_{train}\left(f_\alpha(w)\right).
\end{equation}

From an application perspective, NAS has shown promise in medical imaging, delivering notable performance improvements \cite{benmeziane2024medical}. While improving architectural design, NAS does not inherently offer weight initialization or facilitate feature reuse for fast adaptation. 
Therefore, recent strategies have adopted Supernetworks \cite{cai2019once}, which are over-parameterized neural networks that contain multiple smaller subnetworks within their structure. However, these Supernetworks are task-specific and require retraining when applied to a new target dataset \cite{white2023neural}. 

Addressing both the architecture selection and weight initialization challenges,  Neural Network Search (NNS) has emerged, it leverages a model zoo $(\mathcal{Z})$ of pretrained networks to jointly discover both the architecture $(\alpha)$ and its corresponding weights  $(w)$ based on its alignment with a given Dataset $D$ \cite{jeong2021task}. NNS can be expressed as:
\begin{equation}
(\alpha^*, w^*) = \arg\min_{(\alpha, w) \in \mathcal{Z}} \mathcal{L}^{(D)}_{val}\left(f_\alpha(w)\right).
\label{eq:nns}
\end{equation}

Formulating NNS as a meta-learning problem,
Task-Adaptive Neural Network Search (TANS) \cite{jeong2021task} constructs a cross-modal dataset latent space using contrastive learning to align datasets with high-performing models, thereby enabling efficient retrieval of task-adaptive models. However, TANS faces notable limitations. Specifically, TANS requires training a large number of model-dataset pairs to construct its meta-space, which introduces significant computational complexity with each dataset. This scalability issue makes applying TANS for medical imaging tasks computationally expensive. Additionally, TANS’s reliance on a contrastive loss function emphasizes relationships between individual datasets and models, while overlooking similarities across datasets or architectures. This limitation hinders generalization, which is critical in medical imaging where cross-dataset knowledge transfer is essential.
In this work, we aim to overcome these limitations and extend the approach to the medical tasks; we introduce the first Medical Neural Network Search (MedNNS) methodology. The key contributions of MedNNS include:

\begin{itemize}
    \item We propose MedNNS, the first NNS for medical vision tasks that jointly determines the optimal architecture and pretrained initialization for a given medical dataset by leveraging a pre-constructed meta-learning space.
    \item We leverage Supernetworks to construct a model zoo, enabling efficient subnetwork extraction and expanding the space of model-data pairs.
    \item We introduce a novel rank loss for the meta-space that incorporates inter-model information, ensuring that performance ranking is reflected in model proximity to the dataset representation. 
    \item We introduce the FID loss to capture inter-dataset relationships, ensuring that datasets that have close feature representations are close to each other in the meta-space.
\end{itemize}


\section{Methodology}
\label{sec:method}
\begin{figure}[t]
    \centering
\includegraphics[width=1.\linewidth]{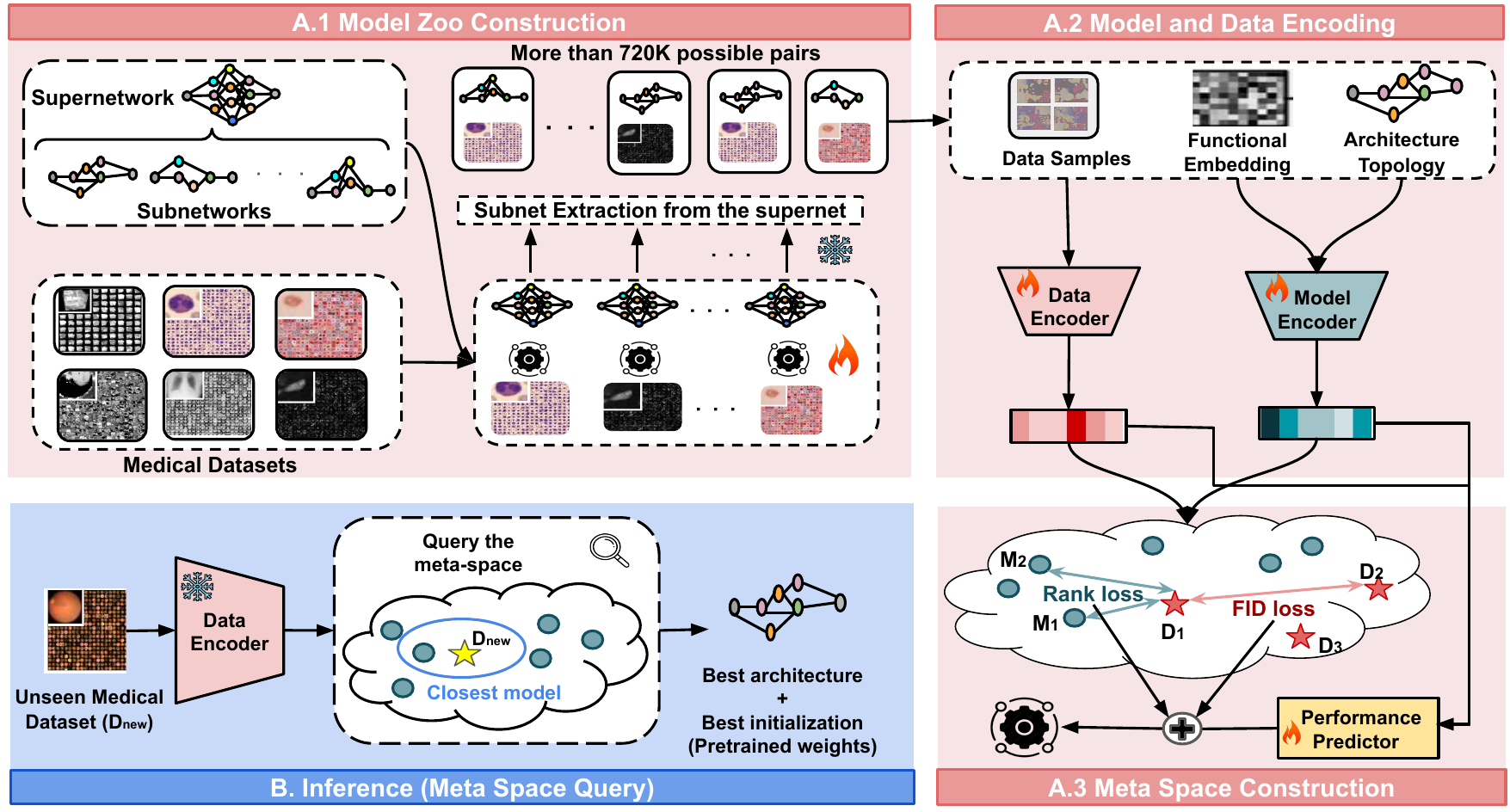}
    \caption{Overview of our MedNNS framework. (A) During training: (A.1) A large model zoo is built by training a single Supernetwork per dataset and extracting thousands of subnetworks via weight sharing. (A.2) Models and datasets are embedded into the latent space. (A.3) The meta-space is optimized using a combination of rank loss, FID loss, and performance loss to align models and datasets according to their relative performance. (B) During inference, given an unseen dataset, its embedding is computed and used to query the meta-space, selecting the closest model embedding as the most suitable pre-trained model.}
    \label{fig:mednns-main}
\end{figure}

We propose MedNNS to solve the NNS problem defined in Eq. (\ref{eq:nns}) by employing Supernetworks to efficiently construct a large model zoo of model-dataset pairs. We also propose a novel approach to embed these pairs into a meta-space that preserves model ranks and dataset similarities. Once built, this meta-space can be queried to select a suitable model and initialization for an unseen target dataset, as shown in Fig. \ref{fig:mednns-main}. In this section, we will describe these steps in detail.

\subsection*{A. Creating the Meta-learning Space}

\textbf{A.1 Model Zoo Construction.} We define a model zoo as $\mathcal{Z} = \{ (D_1, M_1, P_{M_1}^{D_1}), \\ \dots, (D_k, M_l, P_{M_l}^{D_k}) \},$ where $P_{M_l}^{D_k}$ denotes the performance of model $M_l$ on dataset $D_k$. Each model $M$ consists of an architecture $\alpha$ and weights $\theta$, learned by training $M$ on $D$. Constructing an extensive model zoo provides a diverse pool of pairs, enriching the meta-space and boosting overall performance. However, training thousands of models per dataset is computationally prohibitive. TANS \cite{jeong2021task} reduces this cost using accuracy estimation to selectively train model-data pairs, yet it still requires training hundreds of models per dataset. We overcome this inefficiency by training a single Supernetwork per dataset and extracting thousands of subnetworks from it without retraining via weight sharing. Through this, MedNNS enables the efficient construction of a model zoo with 720k model-dataset pairs compared to 14k pairs in TANS \cite{jeong2021task} at a significantly lower computational cost.

\label{sec:A1}
We define a Supernetwork $S(x; \Theta)$ with parameters $\Theta$, while its subnetworks are derived by applying a binary mask $m \in \{0,1\}^n$ to $\Theta$, yielding
$s(x; \theta = m \odot \Theta)$
where $\odot$ denotes elementwise multiplication. Varying masks allow us to extract subnetworks with varied architectures (in depth, width, etc).
Accordingly, the model zoo is represented as
$
\mathcal{Z} = \{ (D_1, s_{1}, \hat{P}_{s_{1}}^{D_1}), \dots, (D_k, s_{l}, \hat{P}_{s_{l}}^{D_k}) \},
$
where each subnetwork $s_l$ is characterized by an architecture $\alpha_l$ and weights $\theta_l$ extracted from $\Theta$ using mask $m_l$. While, $\hat{P}_{s_{l}}^{D_k}$ is the estimated performance of $s_l$ with inherited weights $\theta_l$, and $P_{s_{l}}^{D_k}$ is the true performance when trained from scratch. To ensure that the extracted weights are meaningful, the Supernetwork is trained using a two-stage scheme. Initially, the full network is trained to establish robust shared weights. In later epochs, several subnetworks are sampled each batch, and weights are updated according to the average loss. Also, to ensure rank preservation, which means if
$
P_{s_1} > P_{s_2},
$
then
$
\hat{P}_{s_1} > \hat{P}_{s_2}
$, we employ the FaiRNAS technique \cite{chu2021fairnas} that applies the strict fairness principle during sampling of subnetworks in the second stage of training. Our experiments confirm the effectiveness of this training scheme, as it is demonstrated in the experiments section.

\noindent\textbf{A.2 Model and Dataset Encoding.} 
 For models, an architectural encoding of $\alpha$ is created by flattening configuration parameters such as depth and width. On the other hand, a functional encoding of $\theta$ is acquired at the penultimate layer when passing a fixed Gaussian noise $z \sim \mathcal{N}(0, I)$ through the model. These two encodings are then concatenated and processed through a Multi-Layer Perceptron (MLP) encoder $E_m$ to yield the final model representation. For datasets, a random set of images is selected for each dataset and embedded using a pretrained model; their embeddings are then averaged to produce the dataset representation that serves as input to the MLP data encoder $E_d$.

\noindent\textbf{A.3 Meta Space Construction.} 
We use the acquired Supernet-based model zoo $\mathcal{Z}$ and construct our meta space by aligning model and dataset embeddings, obtained using the encoders $E_m$ and $E_d$, so that their similarities reflect the true performance relationships. We also employ an MLP as a performance predictor ($\varphi$) which takes the model and data embeddings and output $\hat{y}$ as follows $\hat{y} = \varphi\big(E_m(s), E_d(D)\big)$. Following this, we optimize a composite loss defined as: $
\mathcal{L} = \mathcal{L}_{\text{perf}} + \mathcal{L}_{\text{rank}} + \mathcal{L}_{\text{FID}},
$
where \(\mathcal{L}_{\text{perf}}\) is performance prediction loss computed as the mean squared error between the predicted performance and the estimated one $\hat{P}$.
In addition, we introduce a novel rank loss (\(\mathcal{L}_{\text{rank}}\)) to preserve the true ranking of model performances across datasets in the meta-space. For a given dataset \(D\) and any pair of models \(s_j\) and \(s_k\)  with a true performance difference 
$
\Delta P_{j,k}^D = P_{s_j}^D - P_{s_k}^D,
$
we enforce that the corresponding difference in cosine similarity is positive when $\Delta P_{j,k}^D>0$ via a logistic loss:
\begin{equation}
\mathcal{L}_{\text{rank}} = \frac{1}{|P|} \sum_{\Delta P_{j,k}^D > 0} -\log\Big(\sigma\Big(\beta\ .\big(E_d(D) \cdot E_m(s_j) - E_d(D) \cdot E_m(s_k)\big)\Big)\Big),    
\end{equation}
where \(\sigma\) is the sigmoid function and the scaling factor \(\beta\) adjusts the sensitivity. Moreover,  we propose \(\mathcal{L}_{\text{FID}}\) to capture inter-dataset similarities. This loss aligns dataset embeddings based on the similarity of features indicated by the FID metric. For any two datasets \(D_i\) and \(D_j\) with embeddings \(E_d(D_i)\) and \(E_d(D_j)\), a lower FID implies better similarity, so their embeddings should be closer. This is enforced by weighting the squared Euclidean distance between the embeddings:
\begin{equation}
\mathcal{L}_{\text{FID}} = \frac{1}{|P|} \sum_{i \neq j} \exp\left(-\frac{\text{FID}(D_i, D_j)}{\sigma}\right) \|E_d(D_i) - E_d(D_j)\|^2.
\end{equation}

\subsubsection{B. Querying the Meta Space.} During inference, the objective is to select the most suitable architecture $\alpha$ and its initialization $\theta$ for a new unseen target dataset \(D_{new}\) leveraging the pre-constructed meta-space. The latent representation \(E_d(D_{new})\) is computed using the trained dataset encoder \(E_d\) . This representation serves as a query to the meta-space, where a model \(M^*\) is selected by finding the model embedding \(E_m(M^*)\) that is closest to \(E_d(D_{new})\) as follows:
\begin{equation}
M^* = \arg\max_{M \in \mathcal{M}} \frac{E_d(D_{new}) \cdot E_m(M)}{\|E_d(D_{new})\| \, \|E_m(M)\|}
\end{equation}

\section{Experiments}
\label{sec:expiriemnts}
We use MedMNIST \cite{yang2023medmnist} datasets, which comprise a diverse set of medical imaging datasets. Our experimental protocol follows a cross-validation strategy. For each test on a dataset from Table \ref{tab:res}, we construct the meta-space using all remaining datasets from the collection. This approach ensures that the model selection process is unbiased and generalizes well to unseen data. For MedNNS, we employ the OFA (Once For All) Supernetwork \cite{cai2019once}, a ResNet-like model with variable depth, width, and expansion ratio. All Supernetworks are trained on one A100 GPU following the scheme in subsection A1 (\ref{sec:A1}). After that, we evaluate the Spearman rank correlation of subnetworks and achieve 90\% with 0.3\% margin. The meta space is trained using Adam \cite{kingma2014adam} with a learning rate of $10^{-2}$.

For MedNNS, we report three variants to explore the trade-off between efficiency and performance. 1) MedNNS$_{T1}$ is obtained by querying the meta-space based on the target dataset to select the best model and train it. 2) MedNNS$_{T5}$ extends this approach by selecting the top five candidate models, training each for one epoch, and then choosing the model with the highest evaluation accuracy to continue training. 3) MedNNS$_{T10}$ is obtained by extracting the top ten models and then using the same selection process. We compare our approach with SOTA NAS methodologies and DL models initialized using ImageNet pretrained weights. To avoid bias from the choice of optimization strategy, all models are trained using five different strategies, and we report the highest test accuracy achieved at epoch 10 and epoch 100 to reflect both convergence speed and final performance. However, due to the absence of their model zoo structure file, we were unable to test TANS \cite{jeong2021task} method. 

\begin{table}[t]
\caption{Comparison of methods across datasets reporting test accuracies at Epoch~10 (@10) and Epoch~100 (@100) to show convergence speed and final accuracy.}
\label{tab:res}
\centering
\begin{adjustbox}{max width=1.\textwidth}
\begin{tabular}{|l| lcc | c c | c c | c c | c c | c c | c c|}
\toprule
 & \textbf{ Method} 
  & \multicolumn{2}{c}{\textbf{Pneumonia}} 
  & \multicolumn{2}{c}{\textbf{OrganS}} 
  & \multicolumn{2}{c}{\textbf{Tissue}} 
  & \multicolumn{2}{c}{\textbf{Derma}} 
  & \multicolumn{2}{c}{\textbf{Blood}} 
  & \multicolumn{2}{c}{\textbf{Breast}} 
  & \multicolumn{2}{c|}{\textbf{Average}} \\
\cmidrule(lr){3-4}\cmidrule(lr){5-6}\cmidrule(lr){7-8}\cmidrule(lr){9-10}\cmidrule(lr){11-12}\cmidrule(lr){13-14}\cmidrule(lr){15-16}
 & 
  & \textbf{@10} & \textbf{@100} 
  & \textbf{@10} & \textbf{@100} 
  & \textbf{@10} & \textbf{@100} 
  & \textbf{@10} & \textbf{@100} 
  & \textbf{@10} & \textbf{@100} 
  & \textbf{@10} & \textbf{@100} 
  & \textbf{@10} & \textbf{@100} \\
\midrule
\multirow{6}{*}{  \rotatebox{90}{\textbf{DL models                }}  } 
 & MobileNetV3\cite{koonce2021mobilenetv3}     
  & 88.5\% & 92.8\% 
  & 71.3\% & 76.0\%   
  & 58.8\% & 68.1\%   
  & 66.9\% & 75.1\%   
  & 81.0\% & 88.6\% 
  & 73.7\% & 85.9\% 
  & 73.4\% & 81.1\%\\[0.75ex]
 & EfficientNet\textsubscript{B0}\cite{tan2019efficientnet}
  & 85.6\% & 92.3\% 
  & 78.1\% & 81.2\% 
  & 64.4\% & 69.4\%   
  & 70.8\% & 76.2\%   
  & 75.7\% & 87.2\% 
  & 77.6\% & 83.3\% 
  & 75.4\% & 81.6\%\\[0.75ex]
 & EfficientNet\textsubscript{B4}\cite{tan2019efficientnet} 
  & 85.6\% & 92.0\% 
  & 74.5\% & 81.1\% 
  & 55.4\% & 68.7\%   
  & 68.7\% & 74.3\% 
  & 82.8\% & 96.2\% 
  & 73.7\% & 87.2\% 
  & 77.8\% & 83.3\%\\[0.75ex]
 & ResNet18\cite{he2016deep}          
  & 93.1\% & 95.2\% 
  & 66.5\% & 80.6\% 
  & 57.6\% & 66.8\%   
  & 72.0\% & 75.0\%   
  & 91.9\% & 96.2\% 
  & 85.9\% & 89.7\% 
  & 76.5\% & 83.9\%\\[0.75ex]
 & ResNet50\cite{he2016deep}        
  & 91.3\% & 93.9\% 
  & 68.8\% & 81.1\% 
  & 59.9\% & 63.1\% 
  & 72.3\% & 75.0\%   
  & 87.3\% & 92.1\% 
  & 79.5\% & 89.1\% 
  & 80.7\% & 82.4\%\\[0.75ex]
 & DenseNet121\cite{huang2017densely}        
  & 93.6\% & 95.2\% 
  & 79.8\% & 82.1\%
  & 64.0\% & 68.1\% 
  & 72.9\% & 75.9\% 
  & 91.6\% & 96.4\% 
  & 82.0\% & 89.1\% 
  & 80.0\% & 84.5\%\\[1ex]
\midrule
\multirow{6}{*}{   \rotatebox{90}{\textbf{     NAS methods               }}   } 
 & ProxylessNAS\cite{cai2018proxylessnas}   
  & 89.7\% & 94.1\% 
  & 78.3\% & 81.1\% 
  & 63.3\% & 68.6\% 
  & 71.5\% & 76.1\% 
  & 92.4\% & 96.4\% 
  & 84.6\% & 89.7\% 
  & 76.7\% & 84.3\%\\[0.75ex]
 & MNASNet\cite{tan2019mnasnet}           
  & 88.0\% & 93.4\% 
  & 70.4\% & 79.8\% 
  & 63.2\% & \textbf{69.6\%} 
  & 71.5\% & 75.0\% 
  & 88.0\% & 96.2\% 
  & 78.8\% & 87.8\% 
  & 80.17\% & 83.6\%\\[0.75ex]
 & NASNet\cite{zoph2018learning}             
  & 92.9\% & 94.9\% 
  & 77.6\% & 81.9\% 
  & 64.7\% & 68.6\% 
  & 71.0\% & 75.4\% 
  & 91.5\% & 96.0\% 
  & 83.3\% & 90.4\% 
  & 80.17\% & 84.5\%\\[0.75ex]
 & PNASNet\cite{liu2018progressive}             
  & 93.1\% & 94.4\% 
  & 75.8\% & 80.8\% 
  & 63.0\% & 69.1\% 
  & 72.8\% & 75.0\% 
  & 93.3\% & 95.9\% 
  & 82.7\% & 87.2\% 
  & 80.12\% & 83.7\%\\[0.75ex]
 & HardcoreNAS\cite{nayman2021hardcore}             
  & 88.3\% & 92.5\% 
  & 71.5\% & 80.9\% 
  & 62.3\% & 68.7\% 
  & 72.0\% & 75.1\% 
  & 84.8\% & 96.3\% 
  & 79.5\% & 85.3\% 
  & 76.4\% & 83.1\%\\[0.75ex]
 & OFA\cite{cai2019once}                
  & 92.2\% & 93.8\%   
  & 66.4\% & 80.1\%   
  & 58.1\% & 67.2\%   
  & 72.2\% & 75.1\%   
  & 90.3\% & 93.4\%   
  & 84.1\% & 89.3\%   
  & 77.2\% & 83.2\%\\[1ex]
\midrule
\multirow{3}{*}{\rotatebox{90}{\textbf{}}} 
 &  \cellcolor{tmi_blue!10}MedNNS\textsubscript{T1} 
  & \cellcolor{tmi_blue!10}\textbf{93.6\%} & \cellcolor{tmi_blue!10}94.4\% 
  & \cellcolor{tmi_blue!10}79.4\% & \cellcolor{tmi_blue!10}81.8\% 
  & \cellcolor{tmi_blue!10}64.5\% & \cellcolor{tmi_blue!10}69.0\% 
  & \cellcolor{tmi_blue!10}\textbf{75.5\%} & \cellcolor{tmi_blue!10}\textbf{79.7\%} 
  & \cellcolor{tmi_blue!10}\textbf{94.8\%} & \cellcolor{tmi_blue!10}\textbf{96.6\%} 
  & \cellcolor{tmi_blue!10}\textbf{86.5\%} & \cellcolor{tmi_blue!10}\textbf{92.3\%} 
  & \cellcolor{tmi_blue!10}\textbf{82.4\%} & \cellcolor{tmi_blue!10}\textbf{85.6\%}\\[0.75ex]
 &  \cellcolor{tmi_blue!10}MedNNS\textsubscript{T5} 
  & \cellcolor{tmi_blue!10}\textbf{93.6\%} & \cellcolor{tmi_blue!10}\textbf{96.2\%} 
  & \cellcolor{tmi_blue!10}\textbf{80.6\%} & \cellcolor{tmi_blue!10}\textbf{82.1\%} 
  & \cellcolor{tmi_blue!10}64.5\% & \cellcolor{tmi_blue!10}69.1\% 
  & \cellcolor{tmi_blue!10}\textbf{76.2\%} & \cellcolor{tmi_blue!10}\textbf{79.7\%} 
  & \cellcolor{tmi_blue!10}\textbf{95.8\%} & \cellcolor{tmi_blue!10}\textbf{97.5\%} 
  & \cellcolor{tmi_blue!10}\textbf{87.2\%} & \cellcolor{tmi_blue!10}\textbf{92.3\%} 
  & \cellcolor{tmi_blue!10}\textbf{83.0\%} & \cellcolor{tmi_blue!10}\textbf{86.2\%}\\[0.75ex]
 & \cellcolor{tmi_blue!10}MedNNS\textsubscript{T10} 
  & \cellcolor{tmi_blue!10}\textbf{93.8\%} & \cellcolor{tmi_blue!10}\textbf{96.2\%} 
  & \cellcolor{tmi_blue!10}\textbf{80.6\%} & \cellcolor{tmi_blue!10}\textbf{82.4\%} 
  & \cellcolor{tmi_blue!10}\textbf{65.0\%} &  \cellcolor{tmi_blue!10}69.2\% 
  & \cellcolor{tmi_blue!10}\textbf{77.0\%} & \cellcolor{tmi_blue!10}\textbf{79.7\%} 
  & \cellcolor{tmi_blue!10}\textbf{95.8\%} & \cellcolor{tmi_blue!10}\textbf{97.5\%} 
  & \cellcolor{tmi_blue!10}\textbf{90.4\%} & \cellcolor{tmi_blue!10}\textbf{92.3\%} 
  & \cellcolor{tmi_blue!10}\textbf{83.8\%} & \cellcolor{tmi_blue!10}\textbf{86.2\%}\\
\bottomrule
\end{tabular}
\end{adjustbox}
\end{table}

\section{Results and Discussion}
Table \ref{tab:res} demonstrates that our method achieves higher accuracy across all datasets except for the Tissue dataset. Analyzing the selected models, we found a strong correlation between the target dataset and the source dataset used for model initialization, as indicated by the FID scores in Fig.~\ref{fig:fid-finetunning}. For instance, for the OrganS dataset, most of the chosen models are initialized from source datasets such as OrganC and OrganA, which exhibit a good alignment in feature distribution with OrganS. This alignment facilitates feature reuse, which is the key to the observed performance gains. In contrast, the Tissue dataset shows the highest FID difference from the other datasets, indicating a distinct feature distribution that undermines the benefit of reusing features from the meta-space datasets. Moreover, our approach demonstrates remarkably fast convergence: the accuracy at epoch 10 not only surpasses that of competing models at the same stage but, in many cases, matches or even exceeds their final accuracy at epoch 100. This rapid adaptation is a sign of effective feature reuse and it is valuable in medical imaging applications, where data is often limited and efficient learning is crucial.

\begin{figure}[t!]
    \centering
    \includegraphics[width=1.\linewidth]{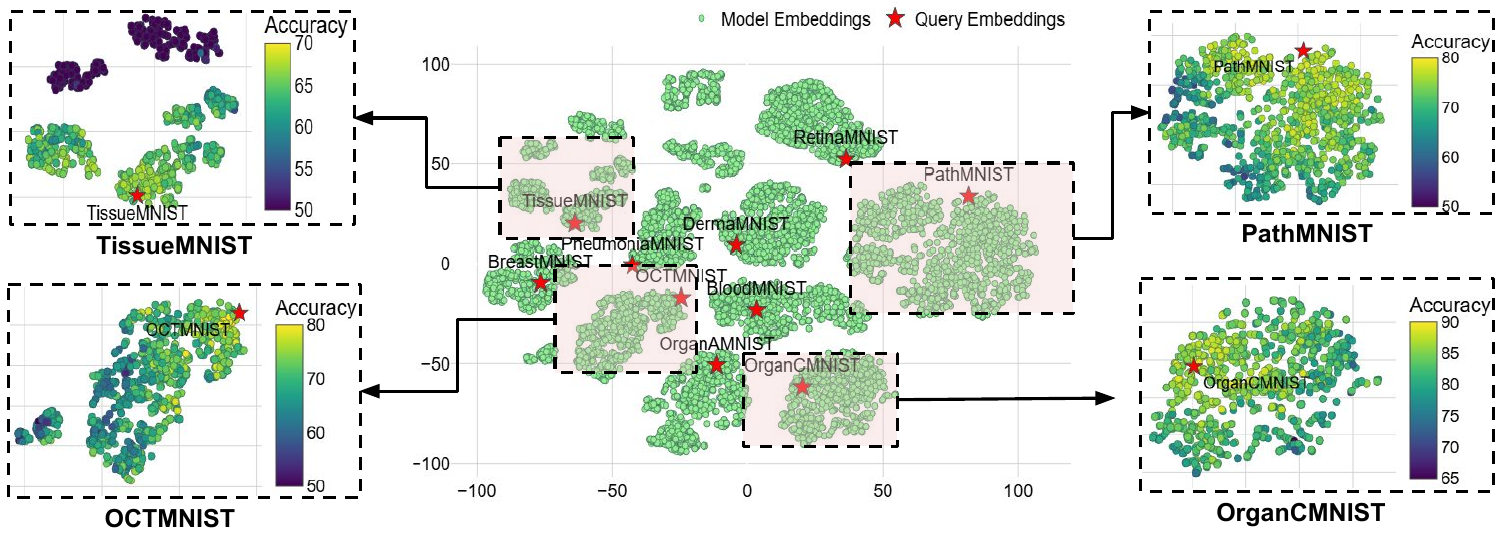}
    \caption{T-SNE visualization of dataset (query) and model embeddings in the meta-space. The central plot shows the MedNNS meta-space. The zoomed-in side plots provide a closer view of specific regions, illustrating the spatial arrangement of models, colored according to their true accuracy, around their corresponding datasets.}
    \label{fig:meta-space}
\end{figure}

\noindent \textbf{Meta-space Analysis.} We analyze the structure of our created meta-space by visualizing the dataset and model embeddings using t-SNE, as shown in Fig. \ref{fig:meta-space}. The central plot reveals that models are clustered based on their performance on datasets, with each group positioned near a corresponding dataset embedding. Also, datasets with similar FID value such as OrganC and OrganA are found in close proximity within the meta-space, validating the efficacy of using the FID loss. To further explore the space, we zoom in on selected regions, as depicted in the four side plots. These visualizations demonstrate a clear correlation between proximity to the dataset embedding and model accuracy: as the distance from the dataset embedding increases, the accuracy decreases. This finding underscores the impact of the rank loss in effectively ordering models by performance.

\noindent \textbf{Ablation.} Table \ref{tab:loss} shows the effect of different loss combinations on constructing the meta-space. Single loss comparison reveals that the rank loss outperforms both contrastive and FID losses. This is because the rank loss enforces the correct order of the models based on performance, while contrastive loss operates at model-dataset pair level without model-to-model interaction. The model ordering attribute of the rank loss can be further observed in Fig. \ref{fig:meta-space}, where model-to-model interaction leads to a continuous meta-space. Pairing losses yields better performance, with the pairing of rank and FID losses delivering the best results.

\begin{table}[t!]
\centering
\caption{Comparison of MedNNS accuracy results using different loss combinations.}
\label{tab:loss}
\begin{adjustbox}{max width=0.6\textwidth}
\begin{tabular}{lcc|ccc|ccc}
\toprule
\multirow{2}{*}{\textbf{Rank}} & \multirow{2}{*}{\textbf{FID}} & \multirow{2}{*}{\textbf{Contrastive}} & \multicolumn{3}{c|}{\textbf{BreastMNIST}} & \multicolumn{3}{c}{\textbf{PneumoniaMNIST}} \\
\cmidrule(lr){4-6} \cmidrule(lr){7-9}
 &  &  & \textbf{T1} & \textbf{T5} & \textbf{T10} & \textbf{T1} & \textbf{T5} & \textbf{T10} \\
\midrule
$\bullet$ & $\circ$ & $\circ$ & 89.8\% & 89.8\% & 91.7\% & 93.0\% & 94.4\% & 94.6\% \\
$\circ$   & $\bullet$ & $\circ$   & 88.5\%     & 90.4\%     & 91.0\%     & 92.6\% & 93.1\% & 95.0\%    \\
$\circ$   & $\circ$   & $\bullet$ & 88.5\% & 89.8\% & 90.4\% & 91.9\% & 93.2\% & 95.4\% \\
$\circ$   & $\bullet$ & $\bullet$ & 89.7\% & 91.7\% & 91.7\% & 94.4\% & 95.5\% & 95.5\% \\
$\bullet$ & $\bullet$ & $\circ$ & \textbf{92.3\%} & \textbf{92.3\%} & \textbf{92.3\%} & \textbf{94.4\%} & \textbf{96.2\%} & \textbf{96.2\%} \\
\bottomrule
\end{tabular}
\end{adjustbox}
\end{table}

\section{Conclusion}
In this work, we introduce MedNNS, the first medical neural network search framework, which simultaneously identifies the optimal architecture and its corresponding pretrained weights for a new dataset by querying a pre-constructed meta-space. Our approach efficiently increases the number of models represented in the meta-space by leveraging a Supernet-based model zoo, while refining the alignment of embeddings by integrating inter-model and inter-dataset performance information through rank and FID losses. This leads to enhanced performance and faster convergence across various medical datasets. Despite these advancements, the challenge of generalization persists, particularly when confronted with datasets that exhibit significant dissimilarity to those within the meta-learning space (e.g., TissueMNIST). In the future, we intend to incorporate a more extensive and diverse array of medical datasets into our framework to expand its coverage. Additionally, we plan to incorporate hardware constraints as an additional dimension in the search process, with the ultimate goal of augmenting the overall applicability and versatility of MedNNS across a broader spectrum of medical imaging scenarios.
\bibliographystyle{splncs04}
\bibliography{main}
\end{document}